\documentclass[10pt,twocolumn,letterpaper]{article}

\usepackage[pagenumbers]{cvpr} % To force page numbers, e.g. for an arXiv version
\usepackage{booktabs}
\usepackage{cuted}
\usepackage{capt-of}  % 用于在非figure环境中使用caption
\usepackage{amsmath}

\usepackage{etoolbox} % 用于修改命令]
\usepackage[utf8]{inputenc}

\BeforeBeginEnvironment{section}{\vspace{-3mm}} % Section 前的间距
\AfterEndEnvironment{section}{\vspace{-1mm}}   % Section 后的间距

\BeforeBeginEnvironment{subsection}{\vspace{-2mm}} % Subsection 前
\AfterEndEnvironment{subsection}{\vspace{-1mm}}  % Subsection 后

\BeforeBeginEnvironment{subsubsection}{\vspace{-2mm}} % Subsubsection 前
\AfterEndEnvironment{subsubsection}{\vspace{-1mm}}  % Subsubsection 后
\usepackage{xspace}
\newcommand{\method}{Sculpt4D\xspace}
\newcommand{\boldstartspace}[1]{\noindent\textbf{#1}}

\definecolor{cvprblue}{rgb}{0.21,0.49,0.74}
\usepackage[pagebackref,breaklinks,colorlinks,allcolors=cvprblue]{hyperref}
\usepackage{microtype}

\def\paperID{10767} % *** Enter the Paper ID here
\def\confName{CVPR}
\def\confYear{2026}

\title{
\method: Generating 4D Shapes via Sparse-Attention Diffusion Transformers
}

\author{
Minghao Yin$^1$
\quad
Wenbo Hu$^2$\footnotemark[2]
\quad
Jiale Xu$^2$
\quad
Ying Shan$^2$
\quad
Kai Han$^1$\footnotemark[3]
\\[0.5em]
$^1$The University of Hong Kong
\quad
$^2$ARC Lab, Tencent PCG
}

\begin{document}
\maketitle

\begin{strip}
    \centering
    \vspace{-4.3em}
    \includegraphics[width=\textwidth]{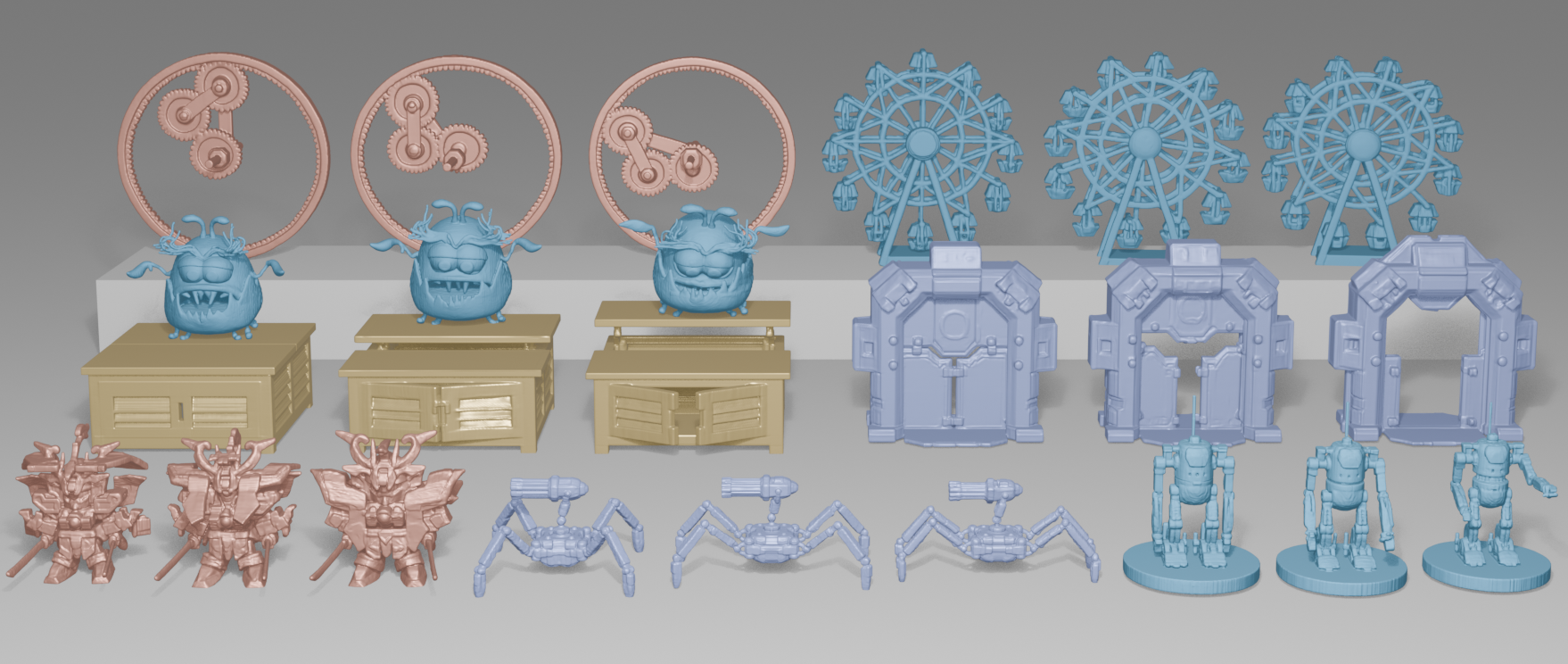}
    \vspace{-0.5em}
    \captionof{figure}{\textbf{High-Fidelity 4D Mesh Generation.} 
    % Sculpt4D generates diverse 4D mesh sequences. We showcase a variety of results, displaying selected keyframes for each sequence to highlight the high-fidelity geometry at different temporal points.
    % \method generates diverse 4D shapes, even for complex motions and topological changes. 
    % Here, we showcase selected keyframes from each sequence to highlight the high-fidelity geometry at different temporal points.
Given input videos, \method generates diverse, temporally coherent 4D mesh sequences, handling complex motions and topological changes. Each row shows selected keyframes from a generated sequence.
%, illustrating the geometric fidelity maintained across time.    
    }
    \label{fig:teaser}
%    \vspace{-1em}
\end{strip}

\begingroup
\renewcommand{\thefootnote}{}
\footnotetext{$^\dagger$ Project lead. \quad $^\ddagger$ Corresponding author.}
\endgroup

\begin{abstract}
% We propose a new 4D generative framework for synthesizing high-fidelity, dynamic 3D content. While recent 3D generative models like Hunyuan3D provide powerful spatial priors, adapting them for 4D generation introduces the prohibitive computational cost of full spatiotemporal attention. Our model addresses this challenge by extending a pretrained 3D Diffusion Transformer (Hunyuan3D 2.1) with newly introduced temporal-spatial attention modules. To ensure both computational efficiency and temporal consistency, these modules are implemented using a structured Sparse Attention strategy. This sparse mechanism drastically reduces cost by selectively focusing computation on high-relevance token pairs, such as a global temporal anchor (the first frame) and spatially-aligned tokens, which is critical for motion tracking. This architectural design allows our network to generate complex and temporally coherent 4D content while remaining computationally tractable. Experiments demonstrate that our 4D model achieves state-of-the-art results on 4D generation benchmarks, producing high-fidelity and temporally consistent dynamic 3D content.
% % 
% We will release our code and model to the community.
% 
% 
Recent breakthroughs in 3D generative modeling have yielded remarkable progress in static shape synthesis, yet high-fidelity dynamic 4D generation remains elusive, hindered by temporal artifacts and prohibitive computational demand.
We present \method, a native 4D generative framework that seamlessly integrates efficient temporal modeling into a pretrained 3D Diffusion Transformer (Hunyuan3D 2.1), thereby mitigating the scarcity of 4D training data.
At its core lies a Block Sparse Attention mechanism that preserves object identity by anchoring to the initial frame while capturing rich motion dynamics via a time-decaying sparse mask.
This design faithfully models complex spatiotemporal dependencies with high fidelity, while sidestepping the quadratic overhead of full attention and reducing network total computation by 56\%.
Consequently, \method establishes a new state-of-the-art in temporally coherent 4D synthesis and charts a path toward efficient and scalable 4D generation.
Project page: \href{https://visual-ai.github.io/sculpt4d}{https://visual-ai.github.io/sculpt4d}
%Project page: \href{https://visual-ai.github.io/sculpt4d}{Click Here}
\end{abstract}

\section{Introduction}
\label{sec:intro}

%The remarkable success of diffusion models has fundamentally transformed generative AI, enabling the creation of high-fidelity content in images [refs], videos [refs], and static 3D objects [refs]. State-of-the-art 3D generators, such as Hunyuan3D [refs], can now produce complex geometries and textures from a single image or text prompt. 

% \previous{
% The generative modeling of static 3D objects has witnessed significant breakthroughs. Large-scale Diffusion Transformers (DiTs)~\cite{peebles2023scalable}, such as Hunyuan3D~\cite{hunyuan3d2025hunyuan3d}, are now capable of synthesizing complex 3D geometries and textures with high fidelity. 
% The natural next frontier in this evolution is 4D generation: the synthesis of dynamic 3D content over time. This task, however, introduces the temporal dimension, compounding the challenges of 3D generation with the need for coherent motion and consistent identity, all while facing a massive increase in computational complexity.
% }

The generative modeling of static 3D objects has witnessed significant breakthroughs.
Large-scale Diffusion Transformers (DiTs)~\cite{peebles2023scalable}, such as Hunyuan3D~\cite{hunyuan3d2025hunyuan3d}, can now synthesize complex 3D geometries and textures from a single image with exceptional fidelity.
In this paper, we tackle a central problem in this domain: \textit{synthesizing high-fidelity 4D assets from videos}. This task poses substantial new challenges — it demands coherent motion and consistent identity, while contending with a dramatic increase in computational complexity and severe data scarcity.

To tackle these challenges, several distinct lines of research have emerged.
Early approaches employed score distillation sampling (SDS)~\cite{poole2023dreamfusion} to optimize a 4D representation~\cite{bahmani20244d,ling2024align,singer2023text,zheng2024unified}, but these methods are notoriously slow and often yield unstable results.
Another popular paradigm uses a two-stage pipeline: first generating multi-view videos, then reconstructing a 4D shape from them~\cite{zhang20244diffusion,xiesv4d,wu2025cat4d,wang20254real,zeng2024stag4d,liang2024diffusion4d}.
While more stable than SDS, such approaches often suffer from error accumulation, where inconsistencies from the video generation stage lead to jittery geometry in the final 4D output.

The scarcity of high-quality 4D datasets~\cite{jiangconsistent4d, deitke2023objaverse, deitke2023objaversenips} has led to a more promising direction: adapting powerful pretrained 3D generative models for 4D synthesis.
This ``3D-to-4D'' adaptation has alleviated some issues, but also revealed critical challenges.
Feed-forward methods like L4GM~\cite{ren2024l4gm} offer fast inference by building on deterministic backbones. However, their reliance on image-based representations limits both geometric fidelity and generalizability.
% , while fast, build on deterministic backbones and image-based representations, which limit their geometric fidelity and generalizability.
%
Other attempts, such as V2M4~\cite{chen2025v2m4}, generate meshes frame-by-frame and use post-process optimization, underscoring the critical need for temporal consistency.
Recently, methods like GVFD~\cite{zhang2025gaussian} demonstrate the value of deforming a canonical 3D shape based on a reference frame by training a separate network.  However, this single-frame approach fails to capture geometric changes in later frames, leading to inaccurate deformations.
%\wbhu{[wbhu: what's the drawback of this approach?]}
%
All these strategies treat the pretrained 3D model as a frozen, black-box component to be managed rather than fundamentally extended.
%

% \previous{
% The most principled and powerful approach, explored by ShapeGen4D~\cite{yenphraphai2025shapegen4d} and our own work, is to create a true, end-to-end 4D generative model by directly adapting a 3D Diffusion Transformer (DiT). This is achieved by inserting new temporal-spatial attention layers into the 3D DiT architecture, allowing the model to learn 4D dependencies natively. While this approach correctly models temporal dependencies, it immediately confronts a severe computational wall: a full spatiotemporal attention, operating on a sequence of $T$ frames with $P$ spatial tokens each, scales quadratically as $O\left((T \times P)^2\right)$. This complexity makes scaling to long or high-resolution 4D content computationally prohibitive.
% }

% We posit that \emph{a more effective solution is to build a scalable 4D generative model by seamlessly integrating efficient temporal modeling into the core generative architecture, Diffusion Transformer (DiT).}
% 
We believe that a more effective solution is to integrate efficient temporal modeling directly into the Diffusion Transformer (DiT) architecture, yielding a scalable end-to-end 4D generative model.
Concurrent to this work, ShapeGen4D~\cite{yenphraphai2025shapegen4d} has explored a similar direction by injecting spatiotemporal attention layers into a 3D DiT, enabling the model to learn 4D dependencies natively.
While this strategy correctly captures temporal relationships, it immediately collides with a severe computational wall: full spatiotemporal attention, operating on $T$ frames with $P$ spatial tokens each, scales quadratically as $\mathcal{O}\left((T \times P)^2\right)$.
This complexity makes the scaling to long or high-fidelity 4D content computationally prohibitive.

In this work, we present \emph{\method}, a native 4D generative model that directly addresses the dual challenges of computational efficiency and temporal consistency.
We build upon the powerful Hunyuan3D 2.1~\cite{hunyuan3d2025hunyuan3d} architecture, inheriting its robust spatial generation capabilities.
At its core, we introduce a novel \emph{Block Sparse Attention} mechanism, designed to capture intricate 4D dependencies with both high fidelity and remarkable efficiency.
This mechanism is founded on two principles tailored for 4D generation.
First, to enforce long-range temporal coherence and prevent identity drift, we adapt the ``attention anchor'' concept from recent large language models~\cite{xiao2024efficient}.
In our design, all tokens across all frames attend to the tokens of the very first frame, establishing a stable global anchor for object identity and appearance throughout the sequence.

Second, to model other temporal relations, we introduce a novel \emph{time-decaying sparse mask}.
This design is informed by the ``spatiotemporal energy decay'' principle observed in works like Radial Attention~\cite{li2025radial} and the computational-density goals of FramePack~\cite{zhang2025packing}.
Distinctively, our solution is specifically tailored for motion-aware 4D generation.
Instead of compressing or dropping tokens, we apply a dynamic sparse mask directly to the attention matrix, which enforces a near-diagonal structure whose stride grows with temporal distance.
This formulation ensures that spatially corresponding tokens maintain attention on each other, preserving object coherence and motion consistency, while efficiently pruning a vast number of uncorrelated token pairs.
In our experiments, we observe that this sparse attention mechanism reduces the total network computation by $56\%$ compared to full attention, without compromising generation quality.

In summary, our contributions are threefold:
In summary, our contributions are threefold: (i) We introduce \method, a novel 4D generative model that extends a powerful pretrained 3D generator (Hunyuan3D 2.1) with spatiotemporal transformer blocks, thereby alleviating the reliance on large-scale 4D training data. (ii) At its core, we propose a sparse attention mechanism that combines an attention anchor with a near-diagonal sparse mask, enabling efficient, motion-aware temporal modeling without compromising generation fidelity. (iii) Extensive experiments demonstrate that our approach achieves state-of-the-art results on 4D generation benchmarks in terms of both geometric quality and temporal consistency.

% \begin{itemize}
%     \item We introduce \method, a novel 4D generative model that architecturally extends a powerful pretrained 3D generator (Hunyuan3D 2.1) with spatiotemporal transformer blocks, thereby circumventing the reliance on scarce 4D training data.
    
%     \item We present a novel sparse attention mechanism that synergizes an attention anchor with a near-diagonal sparse mask, achieving efficient, motion-aware temporal modeling without compromising generation fidelity.
    
%     \item We demonstrate that our approach achieves state-of-the-art results on 4D generation benchmarks, enabling high-fidelity and temporally consistent 4D synthesis.
% \end{itemize}   

\begin{figure*}
    \centering
    \includegraphics[width=\textwidth]{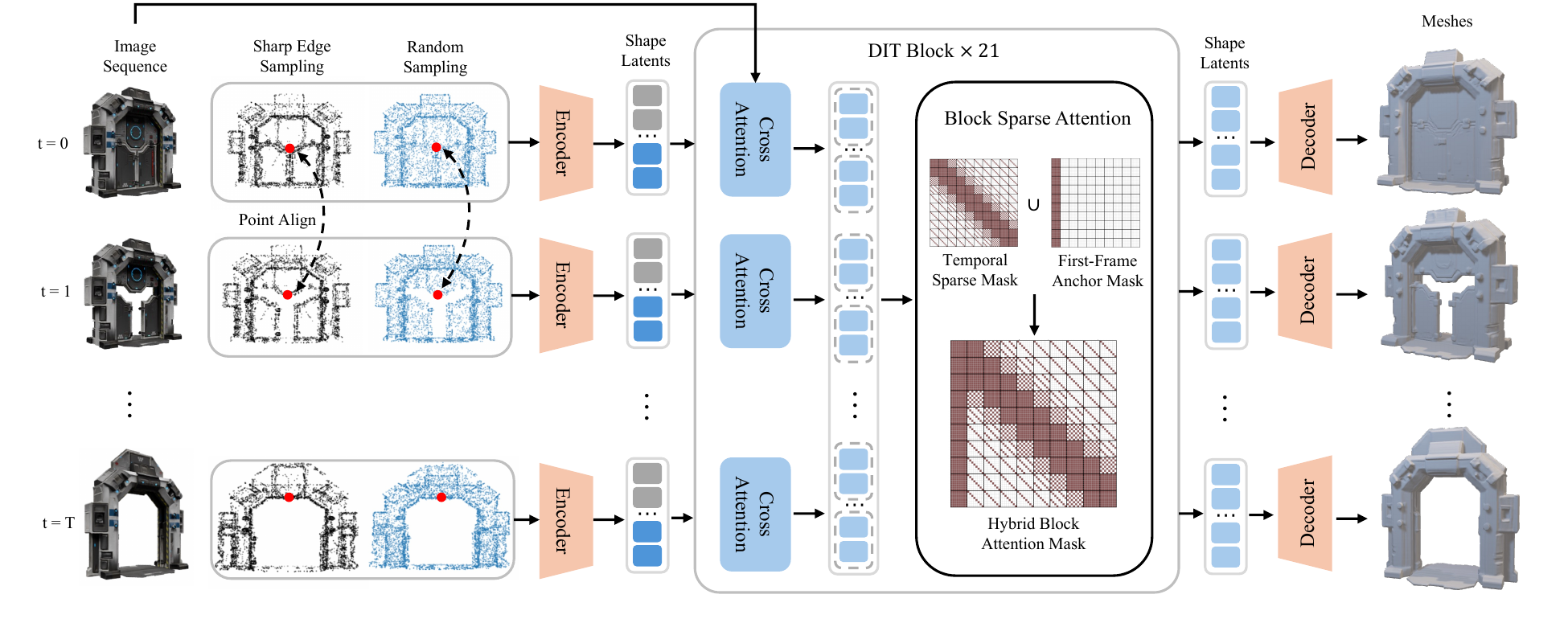}  
    \caption{\textbf{An overview of our 4D generation framework.} Conditioned on an image sequence, we use Consistent Surface Sampling (\cref{sec:consist sample}) to acquire both sharp edge points and random surface points, which a vector set VAE~\cite{zhang20233dshape2vecset, hunyuan3d2025hunyuan3d} encodes into shape latents. These latents are processed by 4D DiT blocks, which use cross-attention for image conditioning and our novel Block Sparse Attention (\cref{sec:sparse attn}). This sparse attention, guided by a composite mask (Temporal Sparse and First-Frame Anchor), efficiently captures motion while ensuring identity consistency. Finally, a decoder produces the final mesh sequence from the denoised latents.}
    \label{fig:framework}
\end{figure*}

\section{Related Work}
\label{sec:related}

% \subsection{3D Generative Models}
\boldstartspace{3D generative models.}
Generative modeling for 3D content has evolved rapidly, moving from early GAN-based~\cite{chan2022efficient, deng2022gram, gao2022get3d, skorokhodov20233d, wu2016learning, xiang2023gram, zheng2022sdf, zhu2018visual} and VAE-based~\cite{caolarge, chen2023primdiffusion, he2024gvgen, jun2023shap, muller2023diffrf, shue20233d, tang2023volumediffusion, wang2023rodin, yariv2024mosaic, zhang2024rodinhd, zhang2024gaussiancube} methods to the now-dominant diffusion models. Initial 3D diffusion works often operated on explicit representations like point clouds or voxel grids. However, a significant breakthrough came with models that generate 3D representations in a compressed latent space, which are then decoded into high-fidelity outputs. This latent-space generation has diverged into two main families: (1) models that generate structured grid or implicit field representations, such as TRELLIS~\cite{xiang2025structured}, and (2) models that tokenize 3D shapes into sets of unordered latent vectors, inspired by 3DShape2VectSet~\cite{zhang20233dshape2vecset}. This second category has led to large-scale, powerful foundation models like Hunyuan3D~\cite{zhao2025hunyuan3d,hunyuan3d2025hunyuan3d} and Step1X-3D~\cite{li2025step1x}, which can generate diverse and high-quality 3D assets from text or images. Our work builds upon this line of research, leveraging a pretrained 3D DiT (Hunyuan3D) as a strong spatial prior for 4D generation.

% \subsection{4D Generation and Reconstruction}
\boldstartspace{4D generation and reconstruction.}
Directly training a generative model for 4D content is exceptionally challenging due to the scarcity of large-scale 4D datasets and the high dimensionality of spatiotemporal data. To circumvent this, the field has developed several distinct paradigms. The earliest and most flexible line of work avoids 4D data entirely, instead using Score Distillation Sampling (SDS)~\cite{poole2023dreamfusion} from pretrained 2D priors. These methods optimize a 4D representation (such as a deformation field or a dynamic NeRF) by enforcing that its rendered views look realistic from all angles and appear consistent over time when guided by a video diffusion model~\cite{ren2023dreamgaussian4d, bahmani20244d, jiangconsistent4d, zhao2023animate124, zheng2024unified, yuan2024gavatar, ling2024align, li2024dreammesh4d, yu20244real}. While versatile, SDS-based approaches are notoriously slow, requiring per-instance optimization, and are often susceptible to artifacts like oversaturation and the Janus (multi-face) problem. 

%A more recent trend aims for feed-forward generation by first producing multi-view videos and then reconstructing a 4D asset from them~\cite{zeng2024stag4d, yangdiffusion, suneg4d, sun2024dimensionx, zhang20244diffusion, jiang2024animate3d, yin2025splat4d, li2024vivid, xiesv4d, yin2025splat4d}. This decouples the problem, but its quality is fundamentally capped by the multi-view video generation step. Inconsistencies or flickering in the generated videos often translate into jittery geometry in the final 4D reconstruction. 

Recent work includes generating multi-view videos for 4D reconstruction~\cite{zeng2024stag4d, yangdiffusion, suneg4d, sun2024dimensionx, zhang20244diffusion, jiang2024animate3d, yin2025splat4d, li2024vivid, xiesv4d}, but quality is capped by the initial video generation, where inconsistencies can cause jittery geometry.

%The most promising direction, which our work follows, is to adapt powerful pretrained 3D generative models to the 4D task. This strategy leverages the rich spatial knowledge learned from vast 3D datasets. However, how this adaptation is performed is the key challenge. A naive approach, seen in V2M4~\cite{chen2025v2m4}, is to use the 3D model as a per-frame generator and then apply a complex, optimization-based post-processing pipeline to register meshes and enforce temporal smoothness. This is not a true end-to-end generative process. A different strategy, employed by GVFD~\cite{zhang2025gaussian}, is to first generate a canonical 3D shape (e.g., from frame 0) and then train a separate diffusion model to predict a deformation field. This method is limited, as it ignores visual information in later frames and struggles to model complex motions or topological changes. Other feed-forward methods like L4GM~\cite{ren2024l4gm} have been proposed, but are often based on non-diffusion architectures and image-based representations, which can limit their geometric fidelity. The most principled approach in this category, explored by ShapeGen4D~\cite{yenphraphai2025shapegen4d} and our work, is to adapt the 3D DiT architecture by inserting new temporal modules, enabling the network to learn 4D dependencies in an end-to-end fashion. The primary bottleneck for this approach is the high computational cost of full spatial-temporal attention, which we address with a novel sparse attention mechanism.
A more promising direction, which we follow, adapts pretrained 3D models to leverage their spatial knowledge. However, the adaptation method is the key challenge. One naive approach, exemplified by V2M4~\cite{chen2025v2m4}, uses a 3D model as a per-frame generator but requires complex, non-end-to-end post-processing to enforce smoothness. Others generate a canonical shape and then predict a deformation field (e.g., GVFD~\cite{zhang2025gaussian}), but this struggles with complex motion and ignores information in later frames. Non-diffusion, image-based methods like L4GM~\cite{ren2024l4gm} can lack geometric fidelity. The most integrated approach, used by ShapeGen4D~\cite{yenphraphai2025shapegen4d} and our work, adapts the 3D DiT architecture with new temporal modules for end-to-end 4D learning. The main bottleneck of this approach is the high cost of full spatiotemporal attention, which we address with a novel sparse attention mechanism.

% \subsection{Efficient and Sparse Attention Mechanisms}
\boldstartspace{Efficient and sparse attention mechanisms.}
The $O(N^2)$ attention bottleneck is a well-known challenge, first addressed in LLMs with structured sparse patterns~\cite{beltagy2020longformer,zaheer2020big} and later with techniques like Native Sparse Attention~\cite{yuan2025native}. This sparsity principle was successfully translated to 3D (Direct3D-S2~\cite{wu2025direct3d} for spatial tokens) and, in parallel, to video generation to handle long temporal sequences.

For video, FramePack~\cite{zhang2025packing} compresses distant frame tokens, while Radial Attention~\cite{li2025radial} applies a static sparse mask with the ``attention sink'' concept introduced in StreamingLLM~\cite{xiao2024efficient}. Our work synthesizes these ideas to manage the computational complexity of our 4D generator. We adopt the ``attention sink'' from Radial Attention as a global anchor for temporal consistency. We pair this with a time-decaying computational density, inspired by both Radial Attention and FramePack. Our novel contribution is the implementation of this decay: a relative diagonal sparse mask that efficiently preserves spatial correspondences over time. This design allows our 4D model to remain computationally tractable while modeling complex, long-range spatiotemporal dependencies.

\section{Method}
\label{sec:method}

\subsection{4D Latent Encoding}
Our goal is to generate 4D dynamic sequences. We build upon the vector set autoencoder~\cite{zhang20233dshape2vecset} framework from Hunyuan3D~\cite{hunyuan3d2025hunyuan3d}, which provides a powerful, compact latent representation for static 3D shapes. Adapting this VAE for dynamic sequences, however, requires a latent space that is not only expressive but also temporally consistent.
A VAE applied naively to individual frames produces temporally jittery latents, which are detrimental to learning smooth temporal dynamics. 
We identify and solve two distinct sources of this jitter: (1) incoherent sampling of the input point cloud on the deforming surface, and (2) stochastic inconsistency from the VAE's reparameterization. We address each in turn below.

% We identify and solve two distinct sources of this jitter: (1) incoherent sampling of the input point cloud on the deforming surface, and (2) stochastic inconsistency from the VAE's reparameterization.

\paragraph{Baseline 3D VAE from Hunyuan3D-v2.1.}

Our autoencoder is built upon the architecture of the Hunyuan3D-v2.1~\cite{hunyuan3d2025hunyuan3d} VAE, which is designed to encode static 3D shapes into a set of latent vectors. The core operation of its encoder involves processing a dense surface point cloud $P$. This point cloud $P$, which includes both uniformly sampled and sharp-edge-focused points for detail, is compressed into a fixed-length representation. This is achieved by first sub-sampling a sparse set of query points $Q$ from $P$ via Farthest Point Sampling (FPS). A cross-attention module, where $Q$ attends to $P$, aggregates the dense point features. Subsequent self-attention layers refine these features into the final latent statistical moments, ($\mu, \sigma$). The decoder, conversely, takes a sampled latent vector $z$ and uses it as a conditioning token to reconstruct an SDF. This baseline architecture provides a strong foundation for 3D representation, but it is inherently designed for static objects and lacks any mechanism for temporal consistency.

\paragraph{Temporally Consistent Surface Sampling.}
\label{sec:consist sample}
When processing a mesh sequence of $T$ frames, $\left\{M_t\right\}_{t=1}^T$, both the encoder's input $P_t$ and its stochastic sampling must be made consistent across time.

The first challenge is that the input point sets are not inherently related. Let $M_t$ be the original deformed mesh for frame $t$. These original meshes $\left\{M_{\text {rest }}, M_1, \ldots, M_T\right\}$ share a consistent topology (i.e., vertex and face indices). However, these $M_t$ are often non-watertight. For robust training, the VAE requires watertight meshes. We generate a watertight mesh $M_t^{\prime}$ for each frame by computing an Unsigned Distance Function (UDF) grid and then applying the Marching Cubes algorithm. Crucially, this process destroys the original topology, meaning $M_t^{\prime}$ and $M_{t+1}^{\prime}$ have completely different and unrelated face/vertex structures.
 
 Naively sampling $P_t^{\prime}$ from $M_t^{\prime}$ at each frame, means the $i$-th point in $P_t^{\prime}$ has no spatial correspondence to the $i$-th point in $P_{t+1}^{\prime}$. It will result in temporally incoherent inputs.
 Our solution bridges this gap by first establishing a canonical mapping. We sample a dense point set $P_{\text {rest }}$ from the canonical rest pose $M_{\text {rest }}$ and store its persistent topological mapping: a face index $f_i$ and its barycentric coordinates $b_i$. Next, we perform temporal propagation for any subsequent deformed frame $t$ by applying this static mapping $\left\{\left(f_i, b_i\right)\right\}$ to the vertices of the corresponding faces on the original deformed mesh $M_t$, which shares the same topology. This yields a ``guide'' point set, $P_{\text {guide, } t}$, which is temporally consistent. Finally, because the VAE must be trained on the clean geometry $M_t^{\prime}$, we project this consistent guide set $P_{\text {guide }, t}$ onto the surface of the final watertight mesh $M_t^{\prime}$. This projection is efficiently approximated by finding the nearest face center on $M_t^{\prime}$ via a k-NN search and adopting its surface position and normal. The resulting point set $P_t^{\prime}$ fed to the encoder is thus both temporally coherent (due to propagation) and on the surface of the high-quality watertight mesh (due to projection).

A consistent input $P_t^{\prime}$ is necessary but not sufficient. The second challenge lies in the VAE's stochasticity. Given the sequence of consistent dense point sets $P_{\text {seq }}^{\prime}=\left\{P_t^{\prime}\right\}_{t=1}^T$, we must also ensure the sparse query points $Q_t$ used by the encoder are consistent. We achieve this by performing Farthest Point Sampling (FPS) only on the first frame's point set $P_1^{\prime}$ to obtain the initial query set $Q_1$. For all subsequent frames $t>1, Q_t$ is defined as the set of points in $P_t^{\prime}$ that correspond to the points in $Q_1$ (tracked via our consistent propagation).
The VAE encoder thus processes a sequence of consistent queries $Q_{\text {seq }}=\left\{Q_t\right\}_{t=1}^T$ and consistent dense points $P_{\text {seq }}$ to produce sequences of per-frame moments: $\mu_{\text {seq }}=\left\{\mu_t\right\}_{t=1}^T$ and $\log \sigma_{\text {seq }}^2=\left\{\log \sigma_t^2\right\}_{t=1}^T$.

%The standard reparameterization trick, $z_t=\mu_t+\sigma_t \cdot \epsilon_t$, introduces an independent noise vector $\epsilon_t \sim \mathcal{N}(0, I)$ for each frame. This stochastic variation severs temporal continuity in the latent space $z_t$, even if the learned $\mu_t$ and $\sigma_t$ are smooth. We modify this sampling process to enforce coherence: instead of sampling $T$ independent noise vectors, we sample a single noise vector $\epsilon_{\text {seq }}$ and broadcast it across the time dimension, applying the exact same stochastic sample to every frame in the sequence.
%The latent representation for any frame $t$ is thus computed as: $z_t = \mu_t + \sigma_t \cdot \epsilon_{seq}$. By sharing the stochastic component $\epsilon_{\text {seq }}$, all random variation is removed from the temporal dimension. Any change in the latent sequence $Z=\left\{z_t\right\}_{t=1}^T$ is now driven only by the learned, deterministic changes in $\mu_t$ and $\sigma_t$, resulting in a temporally-coherent latent space ideal for 4D modeling.

The standard reparameterization $z_t=\mu_t+\sigma_t \cdot \epsilon_t$ samples independent noise $\epsilon_t$ per frame, breaking temporal continuity. To enforce coherence, we sample a single noise vector $\epsilon_{\text {seq }}$ and broadcast it, computing $z_t=\mu_t+\sigma_t \cdot \epsilon_{\text {seq }}$ for all frames. By sharing $\epsilon_{\text {seq }}$, temporal stochasticity is eliminated. The latent sequence $Z$ 's dynamics are now driven purely by deterministic changes in $\mu_t$ and $\sigma_t$, yielding a temporally coherent latent space ideal for 4D modeling.

\subsection{Network Architecture}

Our 4D generation network extends the architecture of the pretrained Hunyuan3D-DiT model~\cite{hunyuan3d2025hunyuan3d}. We adopt this model as our foundation because it provides a powerful spatial feature extractor, trained on a large-scale dataset, which is critical for generalization given the scarcity of high-quality 4D data. Our primary adaptation focuses on introducing spatiotemporal modules to evolve its static 3D generation capabilities into dynamic 4D sequence generation.

To achieve this, we redesign the original DiT block into a new 4D-DiT block. This new block architecture explicitly separates spatial and temporal modeling. Within each block, we first employ the original self-attention and cross-attention module to operate on each frame independently, capturing intricate spatial relationships. Following this, we insert a new temporal self-attention module designed to operate across frames, thereby modeling motion and temporal dependencies. To provide the network with an explicit signal of temporal order, we apply 1D Rotary Position Embeddings (RoPE)~\cite{su2024roformer} along the frame dimension of the query and key tokens in this module.
% to the frame dimension of the query and key tokens processed by this new module.

A critical challenge is the computational complexity of this new temporal module. A naive implementation using full spatiotemporal attention would result in a complexity of $\mathcal{O}\left(N^2\right)=\mathcal{O}\left((T \times P)^2\right)$, where $T$ is the number of frames and $P$ is the number of spatial tokens per frame. This is computationally prohibitive. Therefore, the core of our technical contribution is an efficient block-sparse attention mechanism developed specifically for this temporal module. To ensure training stability, the output projection of this newly added temporal module is zero-initialized, allowing the network to gracefully learn temporal relationships without disrupting the powerful pretrained weights.

\subsection{Sparse Attention Mechanisms}
\label{sec:sparse attn}

Our sparse attention mechanism is guided by two core principles: 1) maintaining long-range temporal fidelity to a reference point, and 2) modeling local temporal relationships with a computationally efficient, decaying density.

%To formally define our mask, we consider a sequence of $T$ frames, where each frame is spatially divided into $P$ token blocks. This 2D structure of ( $T, P$ ) blocks is flattened into a 1D sequence of length $N=T \times P$ for processing. Our temporalspatial attention is then defined by a block-wise attention mask $M \in\{0,1\}^{N \times N}$, where 1 permits attention and 0 denies it.
%To define the mask $M$ at position ( $q, k^{\prime}$ ), we first map these 1D indices back to their 2D temporal-spatial coordinates. A query index $q$ corresponds to frame $i$ and spatial block $k$ via $q=i \cdot P+k$. Similarly, a key index $k^{\prime}$ corresponds to frame $j$ and spatial block $t$ via $k^{\prime}=j \cdot P+t$, where $i, j \in[0, T-1]$ are frame indices and $k, t \in[0, P-1]$ are intra-frame block indices. The value of $M_{q, k^{\prime}}$ is then determined by the following rules.
To define our mask, we first address the sequence structure. The input to the temporal attention module has a total sequence length of $N_{\text {total }}=T \times P$, where $T$ is the number of frames and $P$ is the number of spatial tokens per frame.
%(e.g., 4096, after accounting for the prepended time token).
Our method operates on blocks of tokens to manage this complexity. We define a fixed block size $S_B$ (e.g., 128 tokens). The $P$ spatial tokens of each frame are then spatially divided into $N_B$ contiguous blocks, where $N_B=P / S_B$. %(e.g., given $P=4096$ and $S_B=128$, we get $N_B=32$ blocks per frame).
This 2D structure of ( $T, N_B$ ) blocks is then flattened into a 1D block sequence of length $N_{\text {blocks }}=T \times N_B$. Our spatiotemporal attention is defined by a block-wise attention mask $M \in\{0,1\}^{N_{\text {blocks }} \times N_{\text {blocks }}}$, where 1 permits attention between two blocks and 0 denies it.

To define the mask $M$ at a given query block $q$ and key block $k$, we first map these 1D block indices (where $q, k \in\left[0, N_{\text {blocks }}-1\right]$ ) back to their 2D spatiotemporal coordinates. A query index $q$ corresponds to frame $i$ and spatial block $u$ via $q=i \cdot N_B+u$. Similarly, a key index $k$ corresponds to frame $j$ and spatial block $v$ via $k=j \cdot N_B+v$. Here, $i, j \in[0, T-1]$ are frame indices and $u, v \in\left[0, N_B-1\right]$ are intra-frame block indices. The value of $M_{q, k}$ is then determined as follows.

First, to ensure high fidelity and temporal consistency, we designate the first frame $(j=0)$ as a global anchor. All tokens from all frames $(i \geq 0)$ retain the ability to attend to all tokens within this first frame. This mechanism ensures that the network can always access the initial state, providing a stable reference that prevents semantic drift over long sequences.

Second, for all other temporal relationships ( $j>0$ ), we introduce a time-decaying sparse mask. This design is inspired by the ``spatiotemporal energy decay'' principle observed in works like Radial Attention~\cite{li2025radial}, where token relevance typically decreases with temporal distance. Our approach to reducing computational density for distant frames is conceptually similar to methods like FramePack, which vary patch sizes to achieve a similar effect. Instead of modifying the token representation, our method operates directly on the attention matrix: we retain the full $T \times N_B$ sequence and apply sparsity to the $N_{\text {blocks}} \times N_{\text {blocks}}$ matrix, effectively ``downsampling'' the attention connections rather than the tokens themselves.

The core of this mechanism is a distance-to-stride function, $s(d)$. Given the temporal distance $d=|i-j|$, the function looks up a temporal stride $s$ from a predefined schedule $\mathcal{S}$ (e.g., $\mathcal{S}=[1,1,2,4, 8, 16]$ ): 
$$
s(d)=\mathcal{S}[\min (d, \operatorname{len}(\mathcal{S})-1)]
$$
This stride $s$ defines a relative, diagonal attention pattern. A query token block $u$ is only allowed to attend to a key token block $v$ if they align on this strided grid, i.e., $(u \bmod s(d))=(v \bmod s(d))$.

Combining these two principles, the final block-attention mask $M$ (implemented as an int32 tensor) is formally defined as:
\[
M_{i \cdot N_B + u, \ j \cdot N_B + v} =
\begin{cases}
1, & \text{if } j = 0  \\
1, & \text{if } (u \bmod s_d) = (v \bmod s_d) \\
0, & \text{otherwise}
\end{cases}
\]
The intuition behind the time-decaying sparsity term, $(u \bmod s(d))=(v \bmod s(d))$, is to create a structured, relative sub-sampling of the spatial dimension. This pattern has clear advantages over  a simple ``attend-to-every-s-th-token'' approach (e.g., $v \bmod s(d)=0$ ). This relative pattern ensures that a token block $u$ will always attend to the corresponding token block $v=u$ in a distant frame $j$ (since $u \bmod s=u \bmod s$ is trivially true). This property is vital for maintaining object identity and tracking coherent motion, as it preserves the 1-to-1 spatial correspondence across time. Furthermore, it allows attention to other relative alignments (e.g., $u+1$ attending to $v+1$ ), effectively capturing relative local motion patterns while efficiently skipping a large fraction of uncorrelated spatial pairs, achieving a $1 / s$ computational density.

This formulation has a clear and intuitive effect. For close frames (small $d, s=1$ ), the relation ( $u \bmod 1$ ) = ( $v \bmod 1$ ) is always true, resulting in dense block-attention to capture fine-grained local motion. For distant frames (large $d, s>1$ ), the relation is true for only a fraction $1 / s$ of block pairs, creating a sparse, efficient diagonal-banded attention pattern that respects spatial correspondence. This entire sparse attention mechanism is implemented within our 4D-DIT Block and is executed efficiently using Block Sparse Attention~\cite{guo2024blocksparse}.
\section{Experiments}
\label{sec:experiments}

\begin{figure*}
    \centering
    \includegraphics[width=0.9\textwidth]{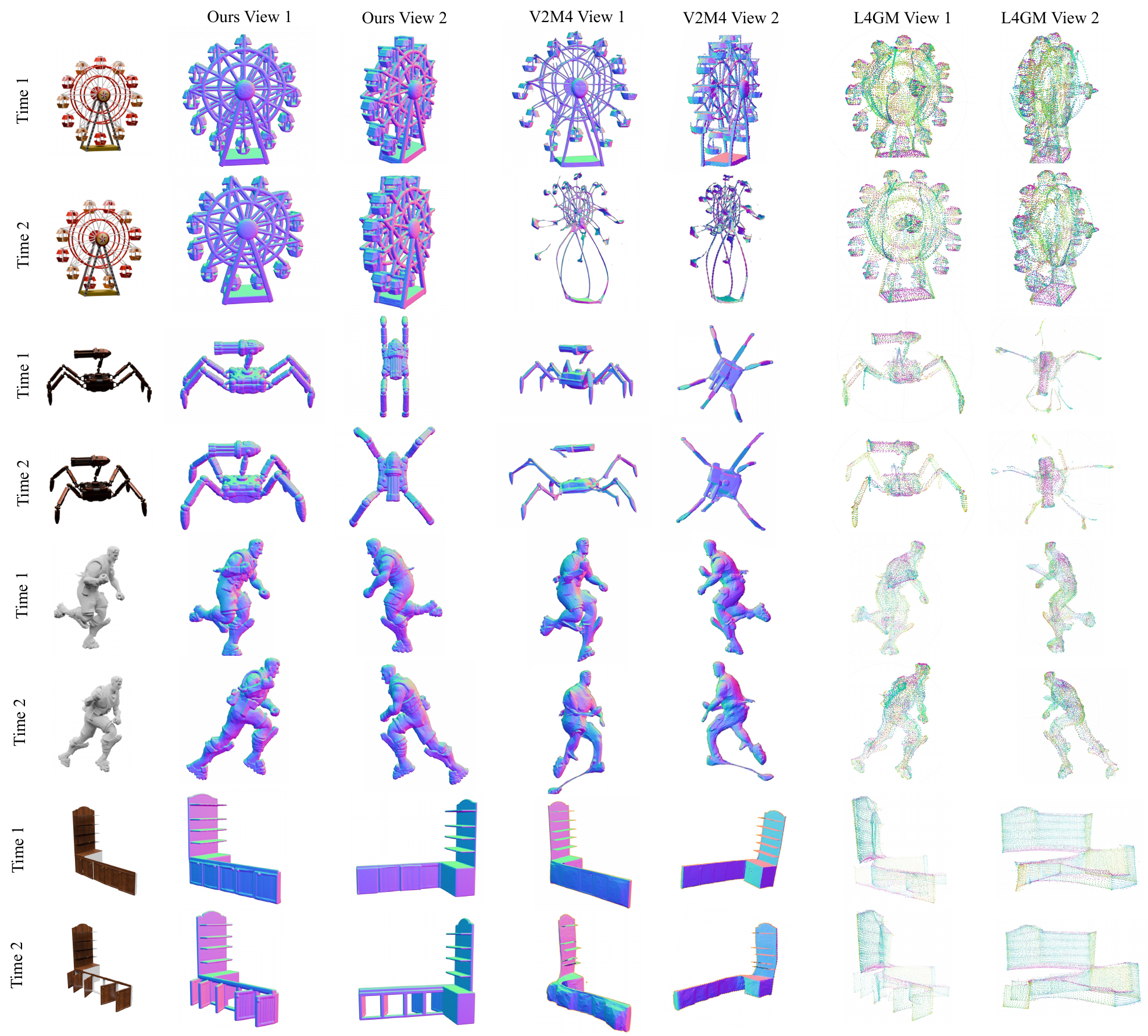}  
%    \vspace{-0.3em}
    \caption{\textbf{Qualitative comparison of 4D mesh generation.} 
    %We compare our method, Sculpt4D, against two baseline methods, V2M4~\cite{chen2025v2m4} and L4GM~\cite{ren2024l4gm}. The leftmost column displays the input image. The subsequent columns show the generated results, with two different views provided for each method. The top and bottom rows correspond to two distinct time frames ($t_1$ and $t_2$ respectively), illustrating the generated geometry at different temporal points. Our method outperforms the baseline methods on visual quality and consistency.}
    We compare Sculpt4D against V2M4~\cite{chen2025v2m4} and L4GM~\cite{ren2024l4gm}. Given an input image (left), we show two generated views per method. Top and bottom rows correspond to time frames Time 1 and Time 2, respectively.}
    \label{fig:comparison}
\end{figure*}

\begin{figure*}[htbp]
    \centering
    \includegraphics[width=0.94\textwidth]{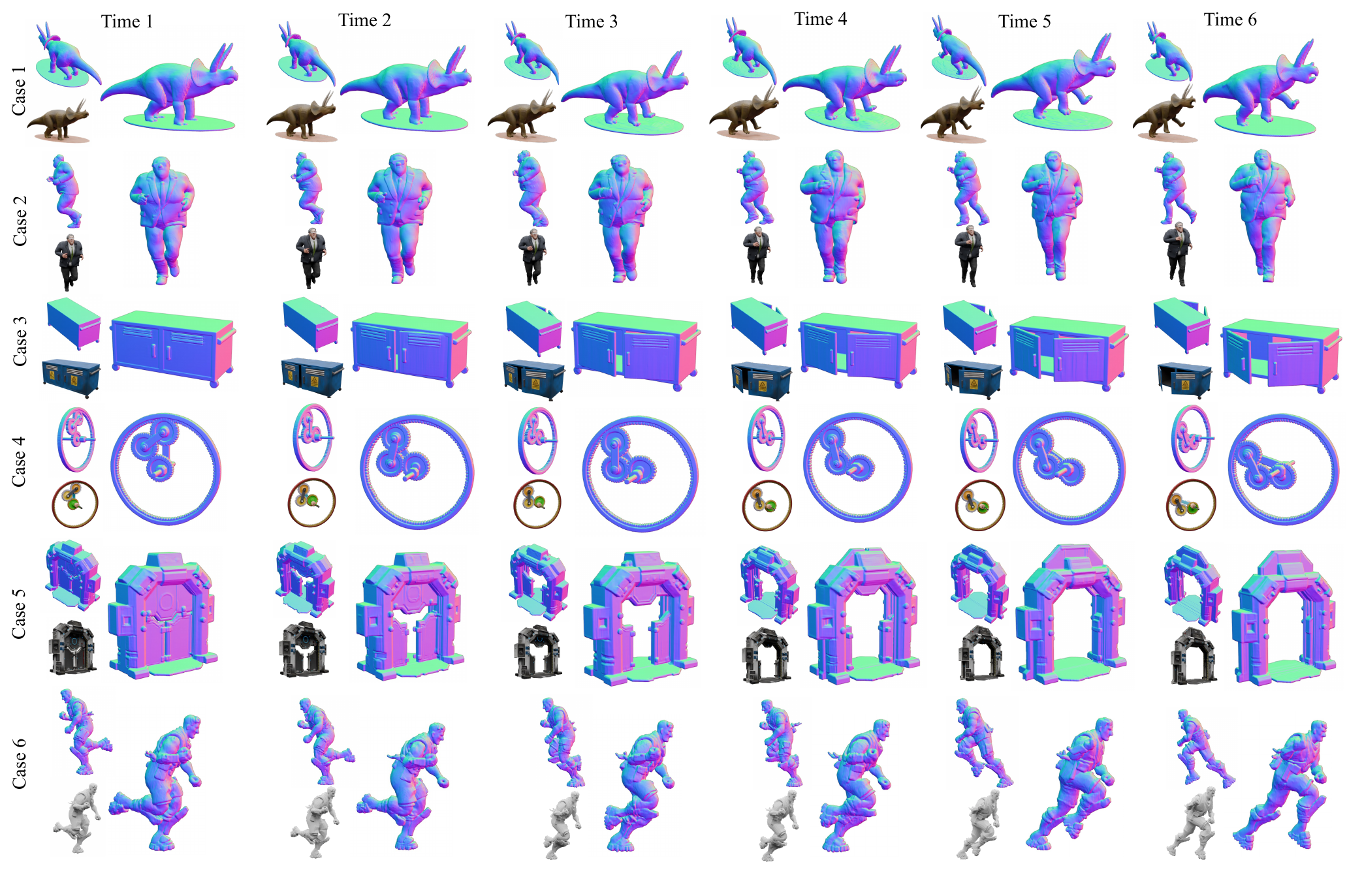}  
    \vspace{-0.3em}
    \caption{\textbf{Additional qualitative results of our method.} 
    %We showcase six diverse 4D generation results (one per row). Each row displays a sequence of six selected time frames, illustrating the temporal progression. For each individual frame, the main image shows the mesh from View 1, while the top-left inset provides View 2 and the bottom-left inset shows the input image.}
    Each row displays one of six diverse 4D results across six time frames. For each frame, the main image shows View 1, with View 2 (top-left) and the input image (bottom-left) as insets.}
    \label{fig:example}
\end{figure*}

\subsection{Data Preprocessing}
We use a training dataset of 13k 4D animated objects filtered from the Objaverse Dataset~\cite{deitke2023objaverse}. Our data preprocessing pipeline, which inherits from the Hunyuan3d-2.1~\cite{hunyuan3d2025hunyuan3d} framework, converts these raw 4D dynamic assets into geometric representations and 2D visual conditions suitable for training. The process is divided into two parallel stages.

First, to provide 2D visual priors, we render multi-view images for each 4D asset at a $512 \times 512$ resolution. For each discrete frame in the animation, we render 24 distinct views. These camera poses are sampled from a sphere using a Hammersley sequence, which is augmented with a randomized offset to ensure varied distributions across different assets. This process results in a 2D image set for each temporal frame, which is paired with the corresponding geometric data for training.

Second, we generate the 4D point cloud representation by Temporally-Consistent Surface Sampling (\cref{sec:consist sample}). To establish strict point-to-point correspondence, we employ a barycentric propagation method. To establish temporal correspondence, we sample a dense set of 124,928 uniform and 124,928 sharp-feature points on the static rest-pose mesh. We then propagate these points through the animation using barycentric coordinates, ensuring point-to-point consistency. To correct for potential mesh artifacts, we project these propagated points onto a clean, high-resolution watertight surface, which is generated for each frame using a flood-fill algorithm on a dense unsigned distance field (UDF). Finally, we aggregate all projected points from the entire animated sequence to compute a single, all-encompassing bounding box. The center and scale derived from this global box are then used to normalize all points, ensuring the entire 4D animation is tightly contained within a $[-1,1]^3$ unit bounding box.

\subsection{Implementation Details}
%We use a pre-trained DINOv2~\cite{oquab2023dinov2} as our image encoder. Input images are resized to 518x518. To ensure consistent conditioning, we process the rendered images as a complete sequence. We first remove the background, then calculate a standard size and center position based on the object's bounding box across all frames. Each frame is resized and centered according to these sequence-level parameters, and the background is filled with white. 
%Within each block, we utilize dimension concatenation to establish skip connections for the latent code, which facilitates robust feature propagation through the deep network.
%Our 4D generation network consists of a stack of 21 4D-DiT blocks. To improve model efficiency and training stability, these blocks incorporate Mixture of Experts (MoE)~\cite{shazeer2017outrageously} layers and utilize RMSNorm~\cite{zhang2019root} for normalization. 
%In Block Sparse Attention, the block size $S_B$ is set to be 128, 1D Rotary Position Embedding (RoPE)~\cite{su2024roformer} is applied to the frame dimension.
%We train the model for 24,000 iterations with a global batch size of 32. Each training sample contains a sequence of 16 frames. 
%During training, we use Farthest Point Sampling (FPS) to select 4,096 query points for loss computation. The entire training process takes approximately 3 days on 8 GPUs with 96GB VRAM.
We extract image features using DINOv2~\cite{oquab2023dinov2} on $518 \times 518$ inputs. For consistent conditioning, we remove backgrounds, then globally align the object's scale and center across the sequence before applying a white background. Our network stacks 21 4D-DiT blocks, utilizing Mixture of Experts (MoE)~\cite{shazeer2017outrageously} and RMSNorm~\cite{zhang2019root} for efficiency and stability. Within each block, concatenation-based skip connections ensure robust feature propagation. We apply 1D RoPE~\cite{su2024roformer} across frames and use Block Sparse Attention ($S_B=128$). Training runs for 24K iterations with a batch size of 32 (16 frames per sequence). Loss computation uses 4,096 query points selected via Farthest Point Sampling (FPS). Training takes $\sim$3 days on eight 96GB GPUs.

\subsection{Main Results}

\paragraph{Quantitative Comparison.}
%As shown in \cref{tab:comparison}, We quantitatively evaluate our model on a dedicated test set comprising 50 4D models from the Objaverse dataset~\cite{deitke2023objaverse}, ensuring that none of these samples were included in the training data. We compare our approach against contemporary state-of-the-art 4D generation methods building on 3D generative models, including L4GM~\cite{ren2024l4gm}, V2M4~\cite{chen2025v2m4} and GVFD~\cite{zhang2025gaussian}.
As shown in \cref{tab:comparison}, we quantitatively evaluate our model on a holdout test set of 50 4D models from Objaverse~\cite{deitke2023objaverse}. We compare against current state-of-the-art methods, including L4GM~\cite{ren2024l4gm}, V2M4~\cite{chen2025v2m4}, and GVFD~\cite{zhang2025gaussian}.
We also establish two reference baselines, Hunyuan3D and Hunyuan3D*, which represent per-frame 3D generation without and with shared noise, respectively. The model is trained on 16-frame input. 
%For our evaluation methodology, we adhere to the protocol established in ShapeGen4D. Since the code for ShapeGen4D~\cite{yenphraphai2025shapegen4d} was not released, we do not include its results in our comparison. This involves assessing the geometric quality of the generated 4D sequences using three primary metrics: Chamfer Distance (CD) and F-Score, which evaluate the similarity between point clouds, and Intersection over Union (IoU), calculated on occupancy voxel grids to measure volumetric overlap. Due to the difficulty in converting the Gaussian particle outputs of L4GM and GVFD into stable watertight meshes, the IoU metric is omitted for these specific methods. As alignment is necessary for a meaningful comparison, we apply the pose registration process defined by this protocol to all generated sequences before metric computation. Our method significantly outperforms all competing methods across all reported metrics, demonstrating its superiority.
We follow the evaluation protocol from ShapeGen4D~\cite{yenphraphai2025shapegen4d}, but omit its results as the code is unreleased. This involves assessing the geometric quality of the generated 4D sequences using three primary metrics: Chamfer Distance (CD) and F-Score, which evaluate the similarity between point clouds, and Intersection over Union (IoU), calculated on occupancy voxel grids to measure volumetric overlap. IoU is omitted for L4GM and GVFD due to difficulties in converting their Gaussian outputs to watertight meshes. 
%Per the protocol, we apply pose registration to all sequences before metric computation. Our method significantly outperforms all competitors on all reported metrics.
As alignment is necessary for a meaningful comparison, we apply the pose registration process defined by this protocol to all generated sequences before metric computation. Our method significantly outperforms all competing methods across all metrics reported, demonstrating its superiority.

\begin{table}[tbp]
\centering
\caption{\textbf{Quantitative comparison.}}
\label{tab:comparison}
\small
\setlength{\tabcolsep}{2pt}
\begin{tabular}{lcccc}
\toprule
 & Representation & Chamfer$\downarrow$ & IoU$\uparrow$ & F-Score$\uparrow$ \\
\midrule
Hunyuan3D & SDF  & 0.1220 & 0.3125 & 0.2820 \\
Hunyuan3D* & SDF  & 0.1231 & 0.3176 & 0.2883 \\
L4GM & MV-3D GS  & 0.1655 & - & 0.2033 \\
V2M4  & mesh + deform  & 0.1268 & 0.3071 & 0.2909 \\
GVFD  & 3D GS + deform  & 0.4235 & - & 0.0717 \\
\textbf{Ours} & \textbf{SDF}  & \textbf{0.1052} & \textbf{0.3381} & \textbf{0.3137} \\
\bottomrule
\end{tabular}
\end{table}

\begin{figure}[htbp]
  \centering
%  \fbox{\rule{0pt}{0.5in} \rule{\linewidth}{0pt}}
  \includegraphics[width=\linewidth]{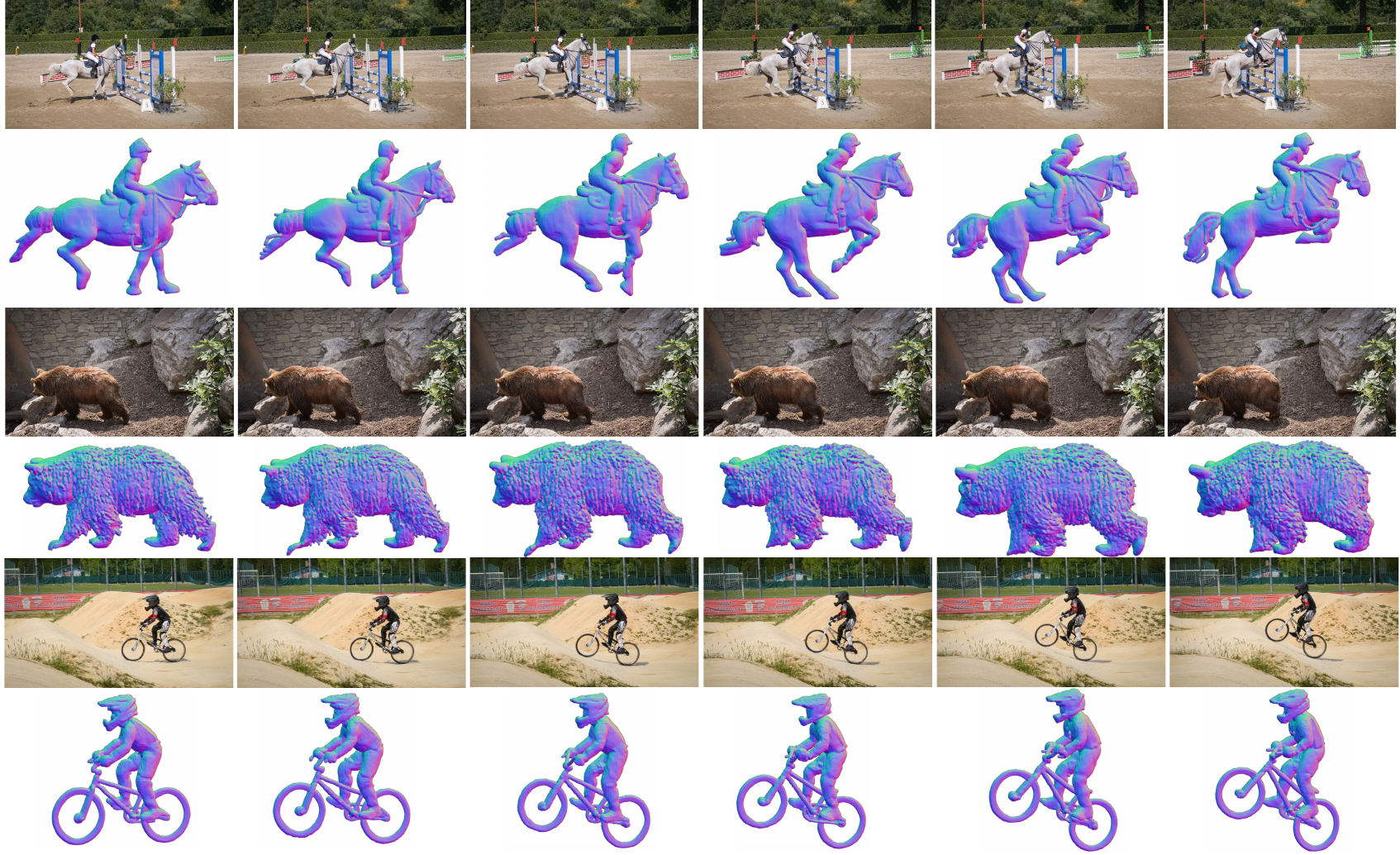}
  \vspace{-0.6em}
   \caption{\textbf{Mesh sequences generated from in the wild data.}}
   \label{fig:inthewild}
   \vspace{0em}
\end{figure}

\begin{figure}[htbp]
  \centering
%  \fbox{\rule{0pt}{0.5in} \rule{\linewidth}{0pt}}
  \includegraphics[width=0.9\linewidth]{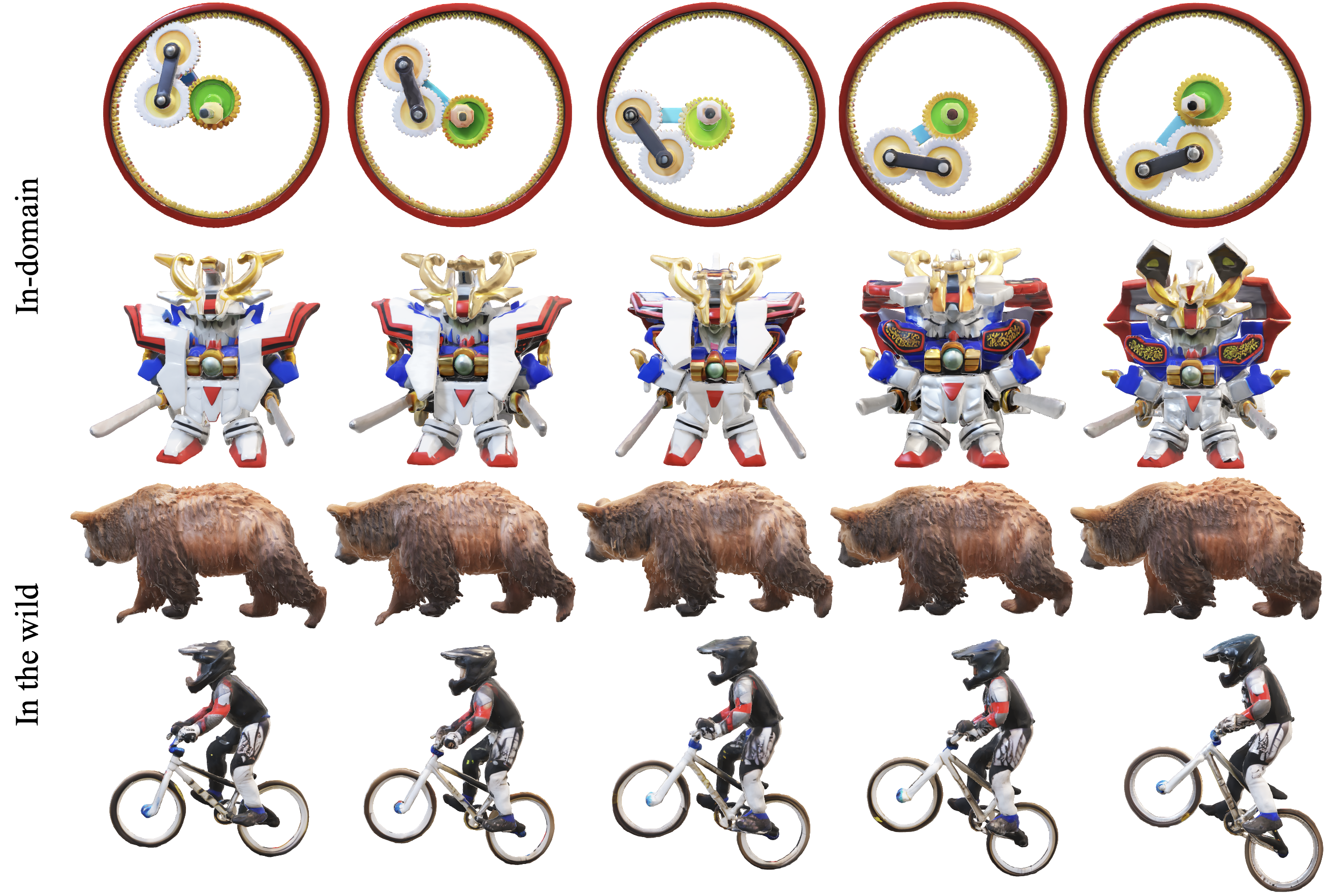}
  \vspace{-0.0em}
   \caption{\textbf{Qualitative results of textured mesh sequences.}}
   \label{fig:textureresults}
   \vspace{-0.0em}
\end{figure}

\paragraph{Qualitative Results.}
As shown in \cref{fig:comparison}, we provide visual comparisons of our model's results against V2M4~\cite{chen2025v2m4} and L4GM~\cite{ren2024l4gm}, showcasing multiple viewpoints and timesteps. In the visualization, the far-left column displays the input video frames. The second and third columns show our generated results, while the fourth and fifth columns present the results from V2M4, and the sixth and seventh columns show those from L4GM. 
Our method produces significantly higher-quality results than the compared methods, with notably better temporal and spatial consistency.

% As can be seen, our method is far superior to the compared methods in terms of overall generation quality, as well as temporal and spatial consistency.

%\paragraph{More Visualizations}
%As shown in \cref{fig:example}, we further demonstrate the capability of our method by presenting a broader gallery of generated 4D objects. We select six representative timesteps from the generated sequences for visualization. In each panel, the bottom-left image is the input video frame, while the main figure on the right displays the generated 3D shape from a primary viewpoint. Additionally, the top-left inlay provides an alternative viewpoint of the same object. These results show that our algorithm consistently maintains high quality and temporal coherence, even when handling complex 4D object dynamics, such as rotating gears, articulated machinery, human motion, and other intricate structures.
As shown in \cref{fig:example}, we present a broader gallery of generated 4D objects across six representative timesteps. In each panel, the bottom-left shows the input frame, the right displays the primary 3D shape, and the top-left inlay provides an alternative viewpoint. These visualizations demonstrate our algorithm's ability to consistently maintain high quality and temporal coherence, even for complex 4D dynamics like rotating gears, articulated machinery, and human motion.

To evaluate the generalization capabilities of Sculpt4D on in-the-wild data, we apply our model to real-world videos from the DAVIS dataset~\cite{perazzi2016benchmark}. As shown in \cref{fig:inthewild}, Sculpt4D robustly generalizes to diverse, unseen dynamics while maintaining high geometric fidelity, demonstrating its effectiveness beyond the training distribution.

To texturize the generated 4D geometry, our framework readily integrates with existing texture pipelines. Adopting a topology-consistent strategy similar to ShapeGen4D~\cite{yenphraphai2025shapegen4d}, we utilize global rigid registration and local ARAP~\cite{sorkine2007rigid} optimization to align sequences to a canonical topology. Consequently, textures generated for the first frame can be seamlessly propagated throughout the entire sequence. \cref{fig:textureresults} showcases our high-fidelity texturing results for both in-domain objects and in-the-wild videos (e.g., a bear).

\begin{table}[tbp]
\centering
\caption{\textbf{Ablation study.}}
\label{tab:ablation}
\small
\setlength{\tabcolsep}{1pt}
\begin{tabular}{lcccc}
\toprule
 & Chamfer$\downarrow$ & IoU$\uparrow$ & F-Score$\uparrow$ & PFLOPs\\
\midrule
w/o consistent sampling  & 0.1128 & 0.3375 & 0.3380  & 186.3\\
w/o shared noise   & 0.1051 & 0.3396 & 0.3342 & 186.3\\
w/o sharp edge sampling   & 0.1005 & 0.3408 & 0.3369 & 186.3\\
w/o attention sink  & 0.0986 & 0.3442 & 0.3375  & 169.8\\
Temporal attention   & 0.2071 & 0.1972 & 0.1833 & 60.2\\
Fixed stride   & 0.1124 & 0.3298 & 0.3306 & 167.1 \\
Full attention    & 0.0958 & 0.3466 & 0.3402 & 425.7\\
\textbf{Ours}   & 0.0972 & 0.3451 & 0.3383 &  186.3 \\
\bottomrule
\end{tabular}
\end{table}

\subsection{Ablation Study}

%As shown in \cref{tab:ablation}, we conduct ablation study to analyze the impact of our key design choices. We evaluate several model variants: (1) w/o consistent sampling, which utilizes random sampling on the watertight surface for each frame's shape VAE encoding instead of our consistent approach; (2) w/o shared noise, where the VAE encoder uses different random noise for each frame rather than shared noise; (3) w/o sharp edge sampling, which is trained using only points from random surface sampling, omitting the sharp edge samples; (4) w/o attention sink, which is trained without first frame anchor design; (5) Temporal attention, which replaces proposed spatial-temporal attention with a temporal-only attention mechanism; and (6) Full attention, which employs a full, non-sparse spatial-temporal attention mechanism instead of our sparse design; (7) Fixed stride, which uses uniform temporal sparsity. The results from these experiments underscore the necessity of our key design components. Notably, our sparse attention design achieves performance that is highly comparable to that of the full spatial-temporal attention, validating its effectiveness. 

As shown in \cref{tab:ablation}, we ablate key design choices across seven variants: (1) w/o consistent sampling, using independent random surface sampling for each frame's VAE encoding; (2) w/o shared noise, applying independent instead of shared noise in the VAE; (3) w/o sharp edge sampling, training only on random surface points; (4) w/o attention sink, removing the first-frame anchor; (5) replacing our spatiotemporal mechanism with temporal-only attention; (6) Fixed stride, applying uniform temporal sparsity; (7) Full attention, using dense instead of sparse attention. The results from these experiments underscore the necessity of our key design components. Notably, our sparse attention design achieves performance highly comparable to that of the full spatiotemporal attention, validating its effectiveness. Additionally, \cref{tab:ablation} evaluates 16-frame computational complexity. Our method significantly reduces the overall load; per layer, sparse attention uses only $35\%$ of the PFLOPs required by full attention, an advantage that scales with sequence length.

%As presented in \cref{tab:ablation}, we evaluate the computational complexity of our model in terms of PFLOPs, calculated using a 16-frame input sequence. The results clearly indicate that our proposed method significantly reduces the network's overall computational load. Delving into a per-layer analysis, our sparse attention mechanism requires only $35\%$ of the computation (PFLOPs) needed by a standard full attention mechanism. This efficiency advantage becomes even more pronounced as the sequence length increases. 
%For example, when the input is extended to 32 frames, our sparse attention layer's computational cost drops to just $21\%$ of that required by full attention. Furthermore, we note that while a simple temporal-only attention module is computationally cheaper, it yields sub-optimal performance because it inherently fails to capture the crucial spatial-temporal relationships necessary for the task.

\section{Conclusion}
\label{sec:conclusion}

We present \method, a native 4D generative framework addressing temporal artifacts and prohibitive computational costs in dynamic synthesis. By integrating efficient temporal modeling, we adapt a pretrained 3D Diffusion Transformer, effectively mitigating 4D training data scarcity. Our core contribution is a novel Block Sparse Attention mechanism. It preserves object identity by anchoring attention to the initial frame, while capturing complex motions via a time-decaying sparse mask. This design faithfully renders spatiotemporal dependencies without the quadratic overhead of full attention, reducing total network computation by $56\%$. Both quantitative and qualitative experiments validate that \method significantly outperforms competing approaches, achieving superior visual quality and temporal consistency.

\section*{Acknowledgments}
This work was supported by the General Research Fund (GRF) of the Hong Kong Research Grants Council (RGC) under Grant No. 17213825.

%In this work, we presented \method, a native 4D generative framework designed to overcome the critical challenges of temporal artifacts and prohibitive computational cost in dynamic synthesis. We successfully adapt a powerful, pretrained 3D Diffusion Transformer by integrating efficient temporal modeling, a strategy that also mitigates the scarcity of 4D training data.
%The core of our contribution is a novel Block Sparse Attention mechanism. This architecture is purpose-built for 4D: it preserves object identity by anchoring attention to the initial frame, while simultaneously capturing complex motion dynamics using a time-decaying sparse mask. This design allows our model to faithfully render complex spatiotemporal dependencies while sidestepping the quadratic overhead of full attention, achieving a $56\%$ reduction in total network computation.
%Our quantitative experiments show that \method significantly outperforms competing approaches, while qualitative results demonstrate superior visual quality and temporal consistency, validating the effectiveness of our method.

% \input{sec/6_finalcopy}

\clearpage
{
    \small
    \bibliographystyle{ieeenat_fullname}
    \bibliography{main}
}

% WARNING: do not forget to delete the supplementary pages from your submission 
\clearpage
\setcounter{page}{1}
\maketitlesupplementary

%\section{Supplementary Webpage}
%We provide \textbf{\texttt{visualization.html}} to display mesh sequences from our 16- and 32-frame models; please allow a few seconds for loading. Note that meshes were decimated to $5\%$ to $10\%$ of the original face count for smooth rendering, resulting in reduced detail compared to the full-resolution evaluation models.
\setcounter{figure}{0}
\setcounter{table}{0}
\renewcommand{\thefigure}{A\arabic{figure}}
\renewcommand{\thetable}{A\arabic{table}}

\section{Model Details}
\label{sec:details}

The network architecture is instantiated as a 21-layer Diffusion Transformer block with a hidden dimension of 2,048 and 16 attention heads, resulting in a head dimension of 128. The model processes a spatiotemporal input sequence generated by the VAE encoder~\cite{zhang20233dshape2vecset}, where each frame consists of 4,096 spatial tokens derived from a 64-channel latent input. Conditioning is provided via cross-attention to visual context embeddings with a dimensionality of 1,370. Within the attention mechanisms, we employ RMSNorm for the normalization of query and key projections to enhance training stability, while standard LayerNorm is utilized for pre-norm blocks.

To facilitate effective information propagation across the network depth, the model incorporates symmetric long-range skip connections between the shallow and deep layers. Specifically, the feature maps from the first 10 layers are cached during the forward pass. The 11th layer acts as a central processing block without skip interactions. Subsequently, the final 10 layers (layers 11 through 20) retrieve the corresponding cached features in reverse order. In these layers, the input hidden states are concatenated along the channel dimension with the retrieved features, temporarily doubling the hidden dimension to 4,096. This concatenated representation is immediately projected back to the standard 2,048 dimensions via a linear transformation and normalized before entering the attention blocks. Within each block, the architecture follows a factorized design that sequentially applies spatial self-attention, cross-attention, and Block Sparse Attention. To scale model capacity, the feed-forward networks in the final six layers are replaced by Mixture-of-Experts (MoE)~\cite{shazeer2017outrageously} modules, featuring eight experts per layer with a top-2 routing strategy.

Temporal positional information is encoded using Rotary Positional Embeddings (RoPE)~\cite{su2024roformer} applied exclusively to the temporal attention layers. We pre-compute sinusoidal frequencies for the frame sequence and apply rotation to the query and key tensors. To match the architectural head dimension of 128, the sine and cosine components—initially computed for half the dimension—are interleaved and duplicated along the last axis. Crucially, to maintain numerical stability during mixed-precision training, this rotational transformation is explicitly cast to and executed in 32-bit floating-point precision before the tensors are reverted to the model’s native data type for subsequent attention computation.

\section{Ablation Study on Attention Mask}

\begin{table}[htbp]
\centering
\caption{\textbf{Ablation study.}}
\label{tab:ablation_sup}
\small
\setlength{\tabcolsep}{1pt}
\begin{tabular}{lcccc}
\toprule
 & Chamfer$\downarrow$ & IoU$\uparrow$ & F-Score$\uparrow$ & PFLOPs\\
\midrule
A: w/o attention sink  & 0.0986 & 0.3442 & 0.3375  & 169.8\\
B: Fixed stride   & 0.1124 & 0.3298 & 0.3306 & 167.1\\
C: Aggressive decay   & 0.0991 & 0.3420 & 0.3365 & 145.0\\
D: Conservative decay   & 0.0968 & 0.3454 & 0.3388 & 233.6\\
E: Full attention    & 0.0958 & 0.3466 & 0.3402 & 425.7\\
F: \textbf{Ours}   & 0.0972 & 0.3451 & 0.3383 &  186.3 \\
\bottomrule
\end{tabular}
\end{table}

%To rigorously validate the effectiveness of our proposed Block Sparse Attention, we conducted comprehensive ablation studies focusing on its two core components: the First-Frame Anchor and the Time-Decaying Sparsity mask. Our experimental design is driven by the trade-off between preserving structural integrity and computational efficiency in long-sequence 4D generation. First, generative models often struggle to maintain consistent geometry over time, leading to structural degradation. The "First-Frame Anchor" is hypothesized to enhance global coherence by providing a constant, ground-truth reference. Second, while modeling temporal dependencies is crucial for motion smoothness, full attention is computationally prohibitive. We hypothesize that video data exhibits a natural decay in information density over time—immediate neighbors require dense attention to capture fine-grained local motion dynamics, whereas distant frames primarily provide semantic context that can be efficiently captured with sparse connections.
To validate our Block Sparse Attention, we conducted ablation studies on its core components: the First-Frame Anchor and the Time-Decaying Sparsity mask. Our design addresses the trade-off between structural integrity and efficiency in 4D generation. First, the ``First-Frame Anchor'' is introduced to mitigate structural degradation by providing a constant reference for global coherence. Second, recognizing that full attention is computationally prohibitive, we hypothesize that information density decays over time: immediate neighbors require dense attention for motion dynamics, whereas distant frames provide semantic context efficiently captured by sparse connections.

To test these hypotheses, we designed two sets of comparisons. Set 1: Effectiveness of First-Frame Anchor. To assess the necessity of the anchor for sustaining global generation quality, we trained a variant (Model A) where the global connection to the first frame is removed, relying solely on the relatively sparse mask. Set 2: Impact of Stride Strategies. To investigate the optimal trade-off between motion smoothness and efficiency, we compared our proposed stride schedule against three representative alternatives: Model B (Fixed stride) applies a constant sparsity (stride=4) across all temporal distances; Model C (Aggressive decay) uses an immediate exponential decay schedule $[1, 2, 4, 8, 16, 32]$; and Model D (Conservative decay) uses a conservative schedule $[1, 1, 2, 2, 4, 4]$. Our proposed Model E (Ours) utilizes a ``Delayed Exponential" schedule $[1, 1, 2, 4, 8, 16]$. We quantitatively evaluated the generation quality using Chamfer Distance, Intersection over Union (IoU), and F-Score. The results, summarized in \cref{tab:ablation_sup}, confirm our hypotheses. Comparing Model A (w/o Anchor) with our full model (Model F), we observe that removing the anchor leads to a degradation in all metrics. This indicates that without the global reference, the model struggles to achieve high geometric fidelity across the sequence. Regarding stride strategies, Model B (Fixed stride) performs the worst, suggesting that uniform sparsity fails to capture essential local motion details. Model C (Aggressive decay) improves efficiency but suffers in geometric accuracy due to the rapid loss of local information. Model D (Conservative decay) achieves competitive results but incurs significantly higher computational overhead. Our proposed Model F achieves the best balance, delivering high IoU and F-Score comparable to the dense Step Decay strategy while maintaining the efficiency benefits of exponential sparsity. This confirms that our hybrid strategy—maintaining dense local attention for generation quality while using exponential sparsity for long-range dependencies—is optimal for 4D generation.

\section{Computational Analysis}

\begin{table}[htbp]
\centering
\caption{\textbf{Computational analysis.}}
\label{tab:comp_sup}
\small
\setlength{\tabcolsep}{1pt}
\begin{tabular}{lcccc}
\toprule
 Frames &  PFLOPs$_{sparse}$ & PFLOPs$_{full}$ & $\frac{Sparse}{Full}$ & $\frac{Sparse\_attn}{Full\_attn} $\\
\midrule
8  & 84.5 & 123.2 & 68.6$\%$  & 58.1$\%$\\
16   & 186.3 & 425.7 & 43.8$\%$ & 35.2$\%$\\
32   & 425.0 & 1584.9 & 26.8$\%$ & 21.5$\%$\\
\bottomrule
\end{tabular}
\end{table}

\begin{figure}[htbp]
    \centering
    \includegraphics[width=0.45\textwidth]{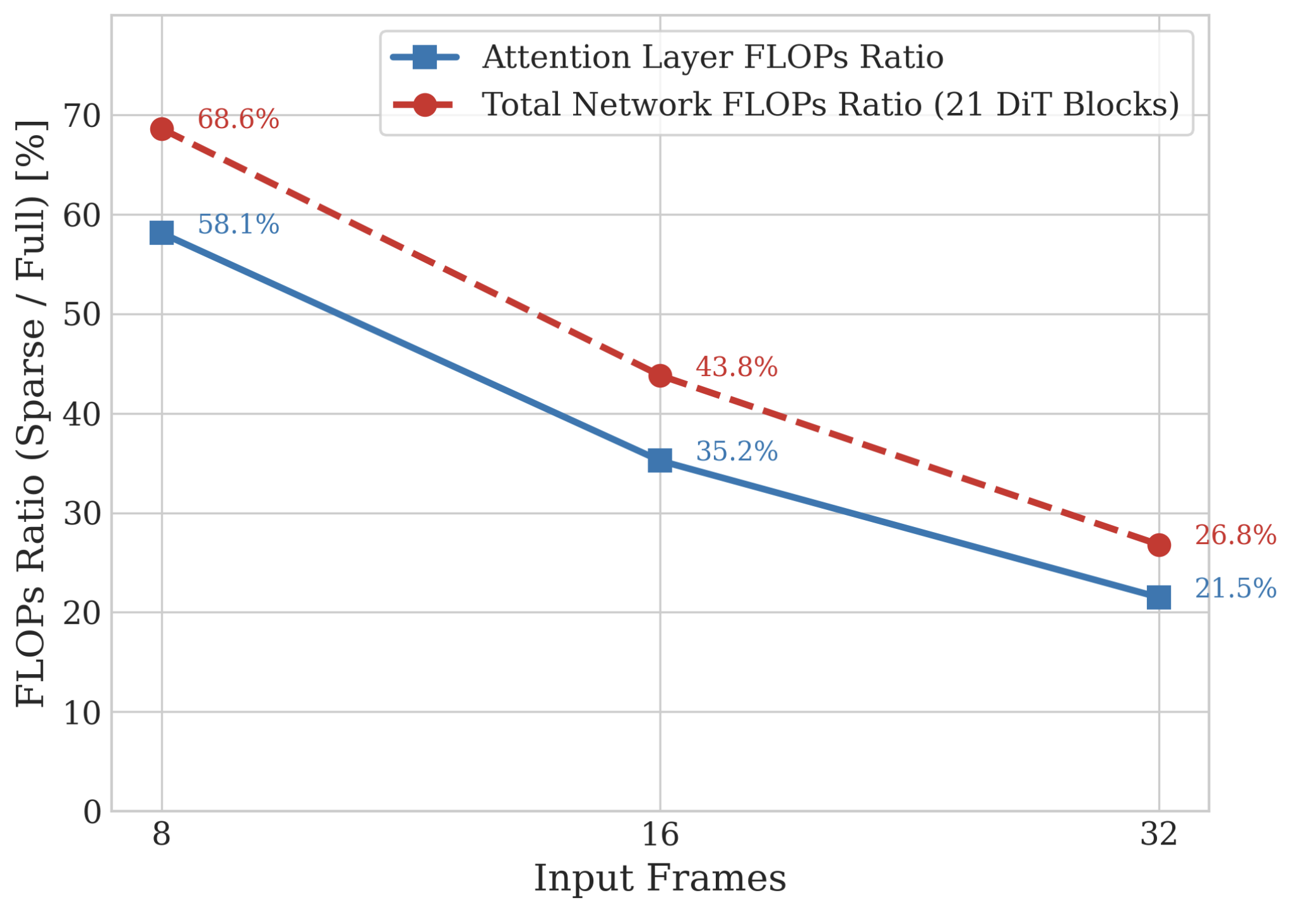}  
    \caption{\textbf{Computational scaling analysis of the sparse temporal attention mechanism}. The lines show the FLOPs ratio (Sparse/Full) for the core temporal attention layer and the entire network across varying input frame counts.}
    \label{fig:comp_sup}
\end{figure}

As shown in \cref{tab:comp_sup}, we quantitatively evaluate the computational benefits of our proposed sparse temporal attention mechanism compared to a standard full attention baseline across different input sequence lengths. When analyzing the core temporal attention layer in isolation, our method demonstrates a highly favorable scaling trend driven by the block-sparse connectivity: while processing a shorter sequence of 8 frames results in $58.11\%$ of the full attention FLOPs, this ratio drops significantly to $35.23\%$ for 16 frames and further decreases to a remarkable $21.50\%$ when handling long 32-frame sequences. This efficiency gain translates directly to the total network computation across the 21 DiT blocks. The full attention network requires 425.7 PFLOPs for 16 frames, whereas our sparse implementation reduces the total cost to 186.3 PFLOPs, or $43.8\%$ of the baseline. For the most challenging case of 32 frames, the full attention model requires a prohibitive 1584.9 PFLOPs; our sparse design dramatically cuts this down to only 425.0 PFLOPs, achieving a mere $26.8\%$ of the total computational load. These results strongly validate the superior efficiency of our block-sparse approach, particularly as the temporal dimension increases.

\cref{fig:comp_sup} clearly illustrates the effectiveness of our sparse attention strategy as the temporal dimension scales. The two curves, representing the FLOPs ratio for the spatiotemporal attention layer and the total network, both exhibit a sharp non-linear decline as the input frame count increases from 8 to 32. This decreasing ratio confirms that the complexity of our sparse mechanism grows much slower than the $O(N^2)$ complexity of full attention. Crucially, the Total Network FLOPs ratio (the upper curve) is consistently higher than the layer-specific ratio (the lower curve). This difference arises because the total network cost includes computationally fixed components, such as Spatial Attention, Cross-Attention, and the Feed-Forward Network (FFN), which must be executed in both the sparse and full baselines. As the temporal attention becomes more sparse, these fixed costs occupy a larger percentage of the total budget, thus raising the overall Network FLOPs ratio, despite the massive savings achieved at the attention layer itself.

\section{Additional Visual Quality Assessment}

%In \cref{tab:vqa_time}, we provide additional quantitative evaluations of Sculpt4D. To comprehensively assess both the visual quality and temporal coherence of the generated sequences, we report video-based quality assessment scores using CLIP, LPIPS, and Fréchet Video Distance (FVD). These metrics collectively demonstrate that our generated 4D content maintains high spatial fidelity and smooth dynamics. Furthermore, we detail the computational cost of our pipeline. On average, generating a complete sequence requires approximately 7 minutes%on a single NVIDIA A100 GPU
%, demonstrating a practical balance between high-quality 4D generation and computational efficiency.
\begin{table}[h]
    \centering
    \caption{\textbf{Results comparison.}}
    \label{tab:vqa_time}
    \small
%    \vspace{-0.8em}
    
    % 【关键修改】使用 resizebox 强制将表格压入当前宽度
    % \linewidth 指的是当前 minipage 的宽度
%    \resizebox{\linewidth}{!}{
%        \setlength\tabcolsep{2pt} % 在缩放前稍微收紧列距，避免字体缩得太小
        \begin{tabular}{lcccc}
            \toprule
            Method & LPIPS $\downarrow$ & CLIP $\uparrow$ & FVD $\downarrow$& Time $\downarrow$ \\
            \midrule
            Hunyuan3D   & 0.131 &  0.803 & 1276.2  & 24 min \\
            DreamMesh4D & 0.145 & 0.835  & 914.9   & 45 min\\
            V2M4        & 0.152 &  0.827 & 952.0   & 45 min \\
            Ours & 0.098 & 0.916  & 483.1   & \textbf{7 min}  \\
            Ours-full & \textbf{0.094} & \textbf{0.919} & \textbf{477.8} & 16 min \\
            \bottomrule
        \end{tabular}
    
    \label{tab:comparison_reb}

\end{table}

In \cref{tab:vqa_time}, we provide a comprehensive quantitative comparison of our method against several baselines (Hunyuan3D~\cite{hunyuan3d2025hunyuan3d}, DreamMesh4D~\cite{li2024dreammesh4d}, and V2M4~\cite{chen2025v2m4}), focusing on video-based quality assessment and overall inference time. To rigorously evaluate the generated 4D sequences, we employ LPIPS for perceptual distance, CLIP score for visual fidelity, and Fréchet Video Distance (FVD) for temporal coherence.
As shown in the table, our method significantly outperforms all baseline approaches across all quality metrics. Most notably, our framework achieves a substantial reduction in FVD.
Furthermore, we evaluate the computational efficiency of each approach. Our default configuration (``Ours") requires only 7 minutes to generate a complete sequence, establishing a new standard for efficiency compared to existing methods that take up to 45 minutes. For scenarios requiring maximum geometric and visual quality, our "Ours-full" configuration achieves the best overall performance (LPIPS of 0.094, CLIP of 0.919, and FVD of 477.8) with a modest increase in inference time to 16 minutes. This demonstrates that our framework not only delivers state-of-the-art 4D generation quality but also provides a highly practical and flexible trade-off between speed and performance.

\section{Generalization to Longer Sequences}

\begin{table}[h]
    \centering
    \small
    \caption{\textbf{Scalability analysis.}}

    \begin{tabular}{lccc} % 注意列数对应
        \toprule
        Frames & Chamfer $\downarrow$ & IoU $\uparrow$ & F-Score $\uparrow$ \\
        \midrule
        8  & 0.099 & 0.338 & 0.315 \\
        16 & 0.102 & 0.339 & 0.315 \\
        32 & 0.106 & 0.334 & 0.314 \\
        64 & 0.114 & 0.326 & 0.310 \\
        % 如果需要视觉上和左表底端对齐，可以在这里加空行，但用 [t] 对齐标题通常更好看
        \bottomrule
    \end{tabular}
    \label{tab:ablation_reb}

\end{table}
        
To evaluate the temporal scalability of Sculpt4D, we investigate its ability to generate sequences longer than those seen during training. Specifically, while our model is trained exclusively on 16-frame sequences, we conduct inference on extended sequences of up to 64 frames without any additional fine-tuning. \cref{tab:ablation_reb} presents the quantitative geometric evaluation—measuring Chamfer distance, Intersection over Union (IoU), and F-Score—across varying sequence lengths (8, 16, 32, and 64 frames). The results demonstrate remarkably robust performance. Notably, even when extrapolating to 4$\times$ the training length (64 frames), the metrics remain highly stable (e.g., the F-Score only experiences a marginal shift from 0.315 to 0.310). This confirms that our incorporated sparse attention mechanism effectively preserves temporal coherence and geometric fidelity, enabling strong zero-shot generalization to significantly longer temporal contexts.

\section{More Visualization Results}

\begin{figure*}[htbp]
    \centering
    \includegraphics[width=\textwidth]{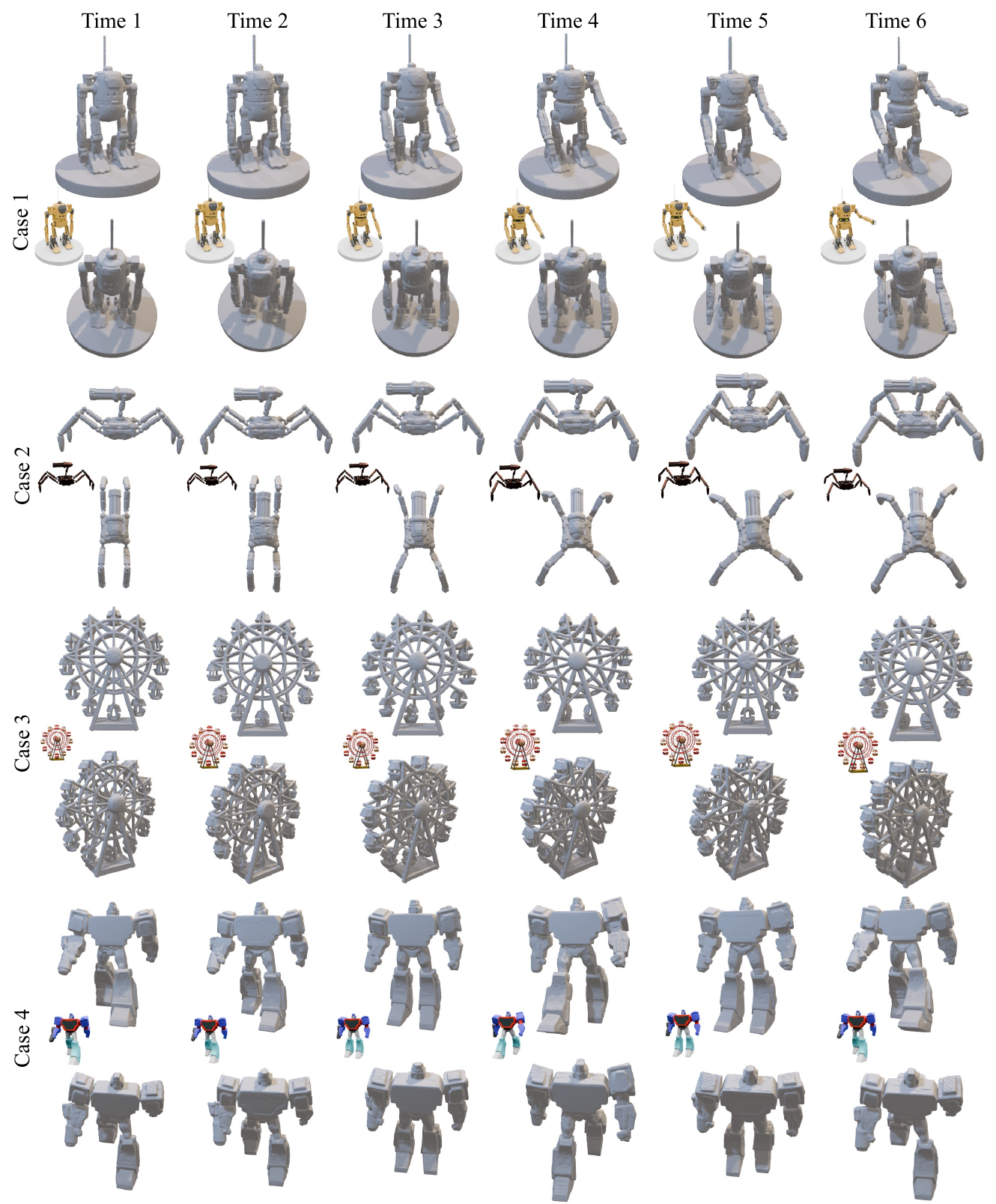}  
    \caption{\textbf{More 4D mesh sequence results.}}
    \label{fig:more1}
\end{figure*}

\begin{figure*}[htbp]
    \centering
    \includegraphics[width=0.95\textwidth]{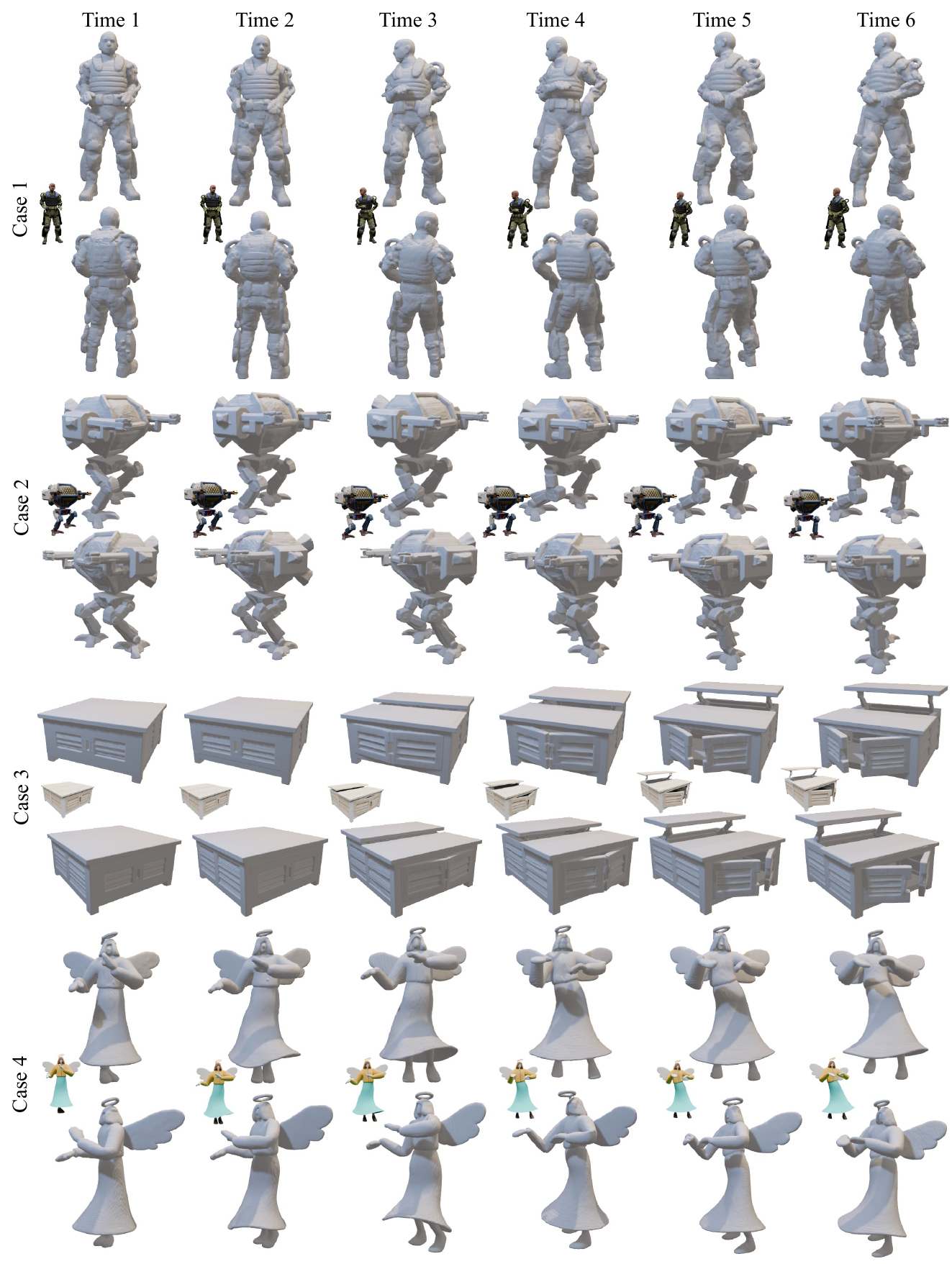}  
    \caption{\textbf{More 4D mesh sequence results.}}
    \label{fig:more2}
\end{figure*}

\cref{fig:more1} and \cref{fig:more2} present additional visualizations of the mesh sequences. We select six time frames and show two views for each frame, with the small images on the left corresponding to the input views.

\end{document}